**World Scientific**
www.worldscientific.com

# Medical X-Ray Image Enhancement Using Global Contrast-Limited Adaptive Histogram Equalization

Sohrab Namazi Nia[*] and Frank Y. Shih[*,†,‡]

*Department of Computer Science, New Jersey
Institute of Technology Newark, NJ 07102, USA*

†*Department of Computer Science and Information Engineering
Asia University, Taichung 413, Taiwan*
‡*shih@njit.edu*

In medical imaging, accurate diagnosis heavily relies on effective image enhancement techniques, particularly for X-ray images. Existing methods often suffer from various challenges such as sacrificing global image characteristics over local image characteristics or vice versa. In this paper, we present a novel approach, called G-CLAHE (Global-Contrast Limited Adaptive Histogram Equalization), which perfectly suits medical imaging with a focus on X-rays. This method adapts from Global Histogram Equalization (GHE) and Contrast Limited Adaptive Histogram Equalization (CLAHE) to take both advantages and avoid weakness to preserve local and global characteristics. Experimental results show that it can significantly improve current state-of-the-art algorithms to effectively address their limitations and enhance the contrast and quality of X-ray images for diagnostic accuracy.

## 1. Introduction

X-ray images have been extensively used in biomedical and medical domains due to the wealth of information they provide for medical diagnosis.[1,11] When an X-ray beam traverses the body, its initial uniform intensity is altered as it interacts with various organs and tissues. The X-ray images have varying levels of brightness due to the different attenuation abilities of various body organs. These tissues can generally be categorized into four distinct groups based on their capacity to attenuate X-rays, as outlined in Table 1. Raw medical X-ray images usually have low contrast, which is difficult for accurate medical diagnosis. Therefore, it is essential to use appropriate image enhancement methods to increase the quality of such important images.

‡ Corresponding author.

Table 1.   Categories of human tissues based on X-ray attenuation ability.

| Brightness Level | Negligible | Low | Middle | High |
|---|---|---|---|---|
| Category | Gas in respiratory, gastrointestinal, and other chamber | Fat | Soft tissues (muscles, solid organ, and body fluids) | Bone and calcium salt |

Various image enhancement and histogram equalization methods have been proposed by researchers.[3,6,7,13] Lavania and Kumar[8] introduced an image enhancement method utilizing filtering techniques with squared masks. However, such methods may inadvertently lose crucial information related to the global characteristics of images due to the use of relatively small masks. Additionally, recent advancements in image enhancement, such as the PLIP-based unsharp masking technique,[18] have emerged to address the limitations of masking techniques and offer improved enhancement capabilities.

Homomorphic filtering[2,4] is another image enhancement algorithm that transfers the image into the frequency domain and applies a homeostasis filter to increase contrast. This method can potentially result in gray score amendments, which might make it not an ideal solution for X-ray image enhancement. Besides, the selection of the most effective homeostasis filter might be a challenging task, specifically for X-ray images. Another TV-homomorphic filtering[16] was proposed to enhance the homomorphic filtering for medical X-ray images.

The Multiscale Retinex algorithm[12] was proposed as an advancement in medical image enhancement. By operating within the HSV (Hue, Saturation, Value) color space, it employs multi-rate sampling to separate color from intensity, enabling comprehensive analysis across various resolutions. By integrating contrast stretching and Multiscale Retinex techniques, it enhances images across multiple scales, and finally all of these scales are combined to reconstruct the final output.

Histogram Equalization (HE), a widely used image enhancement technique, has many different variations based on how it is implemented, but in its most basic form, it maps a gray level from the input image to a gray level in the output. Sukhjinder *et al.*[15] used histogram techniques with median filtering to increase the quality of X-ray images. However, their approach based on median filters could over-smooth the important features of an image. This potential problem can question the generalizability of such a method because edges and some important features might be blurred by mistake using median filters.

Global Histogram Equalization (GHE)[14] used all the pixel values in computing new gray-level values for the pixels. However, the drawback is that it loses the local image characteristics and does not differentiate between closer and further pixels to compute the new gray level for a specific pixel. Additionally, in the absence of an evaluation metric, it is challenging to compare their results against other methods.

Pizer *et al.*[9] proposed Local Histogram Equalization (LHE) to overcome the issues related to GHE, in which the image is divided into tiles, and HE is applied on each tile independently. Although this slicing approach preserves local characteristics, there is a problem of noise-overamplification. To solve this problem, Zuiderveld[19] proposed Contrast Limited Adaptive Histogram Equalization (CLAHE), where a clipping factor is used to control the slope of the Cumulative Distribution Function (CDF) of the histogram. Limiting the slope of the CDF by a clipping factor results in cutting the head of the peak of the histogram and distributing it all over the gray level values. However, since this method works with tiles and only focuses on local image characteristics, it does not consider global image characteristics. Furthermore, the two key parameters of clipping factor and tile size are required to be selected perfectly in order for the algorithm to work correctly.

In this paper, we propose a new approach, named *Global-Contrast Limit Adaptive Histogram Equalization* (G-CLAHE), to take advantage of GHE and CLAHE[10,17] and avoid their drawbacks. The remainder of this paper is organized as follows. Section 2 reviews the GHE. Section 3 describes the Adaptive Histogram Equalization (AHE). We present the proposed G-CLAHE algorithm in Sec. 4. Experimental results are provided in Sec. 5. Finally, conclusions are drawn in Sec. 6.

## 2. Global Histogram Equalization

To enhance image quality based on some pre-defined measure, we need to apply an operator on the input image to obtain the enhanced output. This operator can be conducted on the spatial domain directly or on the frequency domain after image transformation. Fourier transform, wavelet transform, and Discrete Cosine Transform (DCT) are some popular frequency-domain transforms. Figure 1 illustrates the main steps in image enhancement.

HE is widely used in image enhancement to deal directly with spatial domains. It is considered as a mask or global operator depending on the implementation as GHE or AHE. The basic idea is to increase the contrast to make the details of the image more noticeable. The way this algorithm works is by redistributing pixel intensities to achieve a more uniform histogram. This process involves analyzing the histogram of the original image, deriving a transformation to balance the distribution of intensities, and then applying this transformation to restructure the intensity values of the original image. Figure 2 represents a general idea of how HE changes the histogram of an image.

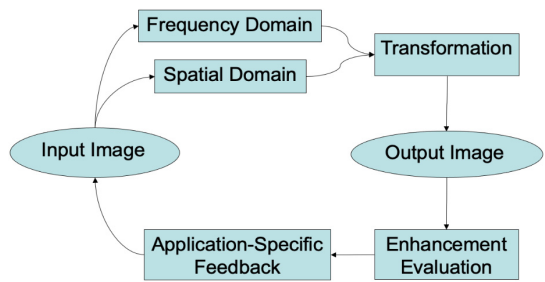

Fig. 1.   Main steps in image enhancement.

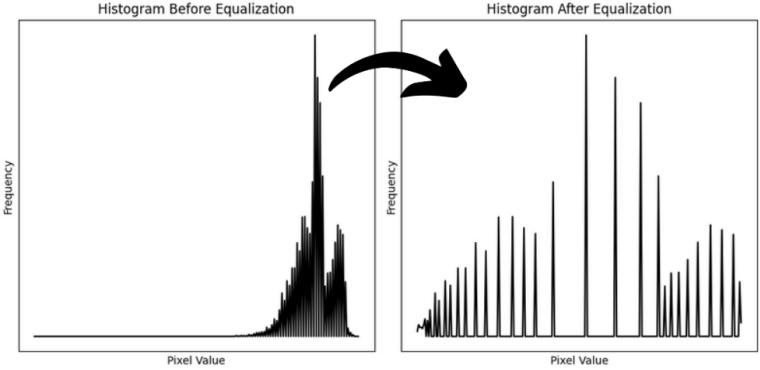

Fig. 2. Effect of HE on an image histogram.

To describe the GHE algorithm, we adopt the notations listed in Table 2.

The GHE algorithm is described as follows:

(1) Compute_Histogram (Image): First, we need to compute the histogram of the input image. It simply means that for each gray level in the range $[0, L-1]$, we count how many pixels with that specific gray level exist in the input image. The histogram values of each gray level are computed as

$$H(i) = \sum_{x=0}^{W-1} \sum_{y=0}^{H-1} \delta(I(x,y) - i), \tag{1}$$

where $\delta$ is the Dirac Delta function in discrete format and $I(x,y)$ denotes the pixel located at $(x,y)$. Hence, we have

$$\sum_{i=0}^{L-1} H(i) = N, \tag{2}$$

where $N$ is the total number of pixels. For example, an image of size $5944 \times 3963$ is shown in Fig. 3 next to its computed histogram.

Table 2. Notations in the global histogram equalization algorithm.

| Symbol | Description |
|--------|-------------|
| $N$ | Total number of pixels of the input image |
| $L$ | Number of unique gray levels ($[0\text{-}L\text{-}1]$) |
| $W$ | Width of the image in terms of pixels |
| $H$ | Height of the image in terms of pixels |
| $\delta$ | Dirac Delta function in discrete form |
| $I$ | The input image |

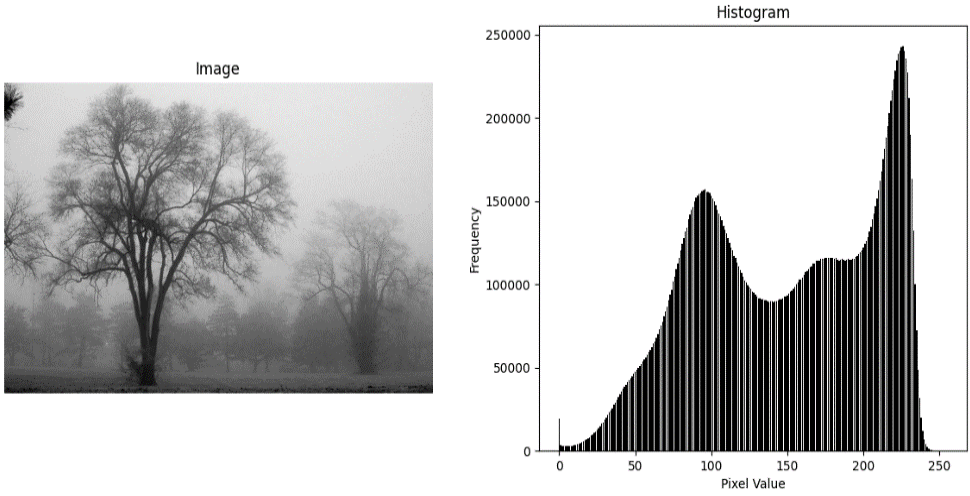

Fig. 3. Histogram of a sample image.

(2) Compute_CDF (Histogram): In this step, we compute the CDF of the histogram as follows. The obtained CDF of the computed histogram in Fig. 3 is shown in Fig. 4.

$$\text{CDF}(i) = \sum_{j=0}^{i} H(j). \tag{3}$$

It is obvious that we have

$$\text{CDF}(L-1) = N. \tag{4}$$

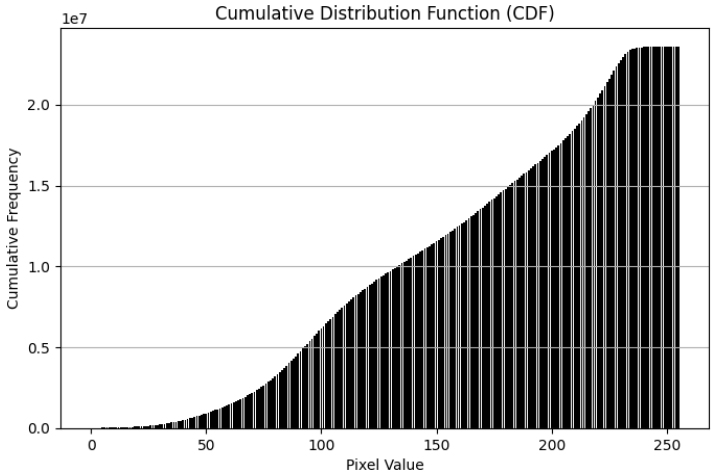

Fig. 4. The CDF of an input image.

(3) Normalize (CDF): In this step, we normalize the CDF by the following equation, so that the new values are in the range of [0, 1].

$$\text{NCDF}(i) = \frac{\text{CDF}(i)}{N}. \tag{5}$$

Hence, we have

$$\text{NCDF}(L-1) = 1. \tag{6}$$

(4) Compute_New_Intensity (NCDF, Image): The new intensity values are obtained by

$$\text{New\_Intensity}(p) = \text{round}(\text{NCDF}[\text{original\_Intensity}(p)] \times 255). \tag{7}$$

The round function simply rounds the floating-point result to the closest integer.

(5) Create_Output(): After computing the new intensity for each pixel, we simply replace the original intensity of each pixel with the new value obtained by Step 4.

Figure 5 shows how HE can practically help to increase the contrast of an image. In this example, there is a secondary peak (shown in a circle) that has significantly lower values as compared to the main peak. Note that this secondary peak may have the potential to convey an important meaning regarding the image, but since the first peak has extremely higher values than that, it is difficult for the secondary peak to be clearly visible in the image. However, after applying HE to make the distribution more uniform, the resulting histogram is now more successful in showing the secondary peak.

Although GHE is a perfect algorithm for increasing image contrast, it only considers global image characteristics and does not differentiate between the neighboring pixels and others. However, in some parts of a medical X-ray image, there might be a case that the local neighboring pixels are more important than the further pixels. Hence, GHE generally loses the local image characteristics.

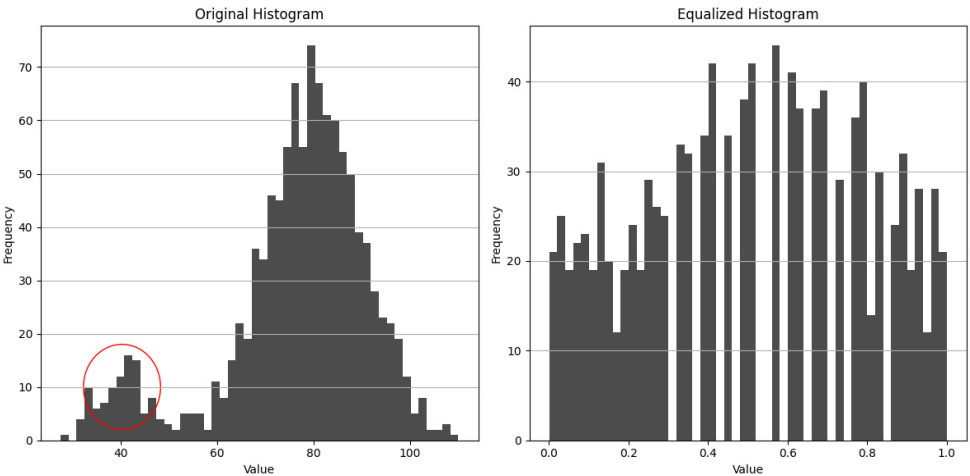

Fig. 5.   Using HE to increase contrast.

## 3. Adaptive Histogram Equalization

The fundamental approach of AHE is to divide an image into smaller regions, which are called tiles, and apply HE independently to each tile. Therefore, in AHE only neighboring pixels influence a pixel's new intensity value. The choice of tile size as a critical parameter is important, and a common practice of $8 \times 8$ is used.

One of the drawbacks of the AHE algorithm is the checkerboard effect since the image is divided into tiles and each tile is processed separately. The resulting pixels around the borders of the tiles might not match perfectly, and therefore, these regions around the borders might look unrealistic. One solution is to use bi-linear interpolation, which can reduce the artificial appearance of the borders of tiles after HE is applied on each tile. It calculates these new pixel values based on a weighted average of the nearest four pixels, resulting in a smoother transition and a more natural-looking result.

However, the AHE can potentially cause noise over-amplification. Assume that while using the AHE algorithm, there is a specific tile that has a noise in it, but apart from that, the rest of the pixel intensities are very uniform. In this case, since HE potentially increases contrast, it can lead to the amplification of noise in that specific tile.

## 4. The Proposed Methodology

In this section, we first introduce the CLAHE, which can avoid the noise over-amplification problem. Noise over-amplification occurs when the AHE extremely increases the contrast of a tile that has noise. The CLAHE controls the slope of the CDF of the image histogram by a clipping factor to avoid too much contrast. If the difference in CDF value of any two adjacent gray levels is greater than the clipping factor, it is set to the clipping factor. With respect to an image histogram, CLAHE cuts the top part of the histogram and then redistributes it all over the histogram. The top part is considered as the histogram values that exceed the clipping factor. The effect of clipping on a sample histogram is shown in Fig. 6.

However, the problem with CLAHE is that it purely focuses on the local image characteristics, and the clipping is carried out on each local region independently. However, there might be some cases in which global image characteristics are as important as local features, and in such cases, CLAHE will have poor performance because it only considers local features, and in some cases, local characteristics might be meaningless. Another challenge in CLAHE is that the clipping factor should be chosen wisely for the algorithm to perform correctly.

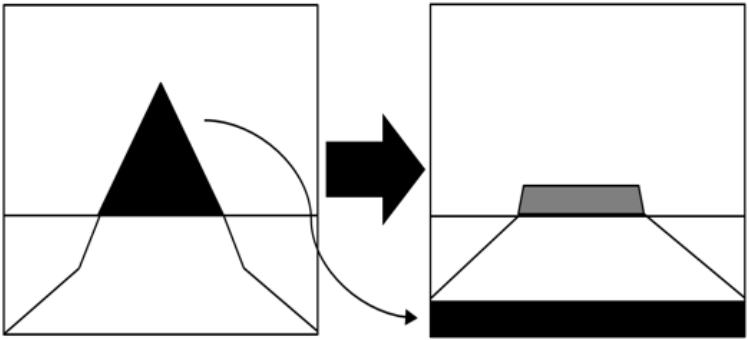

Fig. 6.   Effect of clipping on a sample histogram.

To solve the aforementioned problems, we propose the G-CLAHE to find a balance between preserving local image characteristics and global image characteristics. Unlike traditional evaluation methods based on the original image, our approach assesses local enhancement against the globally equalized image. In more detail, we do not compare the Locally Enhanced Image (LEI) with the original image to evaluate how good or bad the enhancement has performed, but we compare it to the Globally Equalized Image (GEI) of the LEI. Generally speaking, we continue to enhance the image locally over and over in an iteration-based method as long as the LEI goes toward the GEI of itself. When the LEI starts getting more dissimilar from its GEI, it means we have to stop the algorithm. Otherwise, we will lose the image's global characteristics.

Using the globally enhanced version of the image as a reference is a wise decision because the globally enhanced image is a better evaluation reference as compared to the original image. The reason is that the contrast in the globally enhanced image has already been increased, and it has a higher quality as compared to the original image while preserving the global characteristics. Hence, The GHE result with magnified global contrast is a better reference than the initial raw image that might have many potential low-contrast regions.

In the G-CLAHE algorithm, we use a similarity evaluation metric to determine when to terminate the algorithm. Our framework is designed to allow adaptation or selection of any suitable similarity evaluation function. The goal of this iteration-based algorithm is to enhance the input image locally step by step with respect to the globally enhanced image, such that the similarity evaluation metric can be maximized. In our experiments, we adopt various metrics such as SSIM, PSNR, MSE, SCI, RMSE, and MAE. The results demonstrate the effectiveness of the algorithm across all these metrics.

The proposed G-CLAHE algorithm is described below.

First, the GHE of the input image is computed and stored in GEI (globally enhanced image). Let $N$ be the iteration index of the algorithm that is initially set to 0. Tile Size (TS) is a parameter of this algorithm that can be set to an arbitrary

---

**Algorithm 1** G-CLAHE

---

**Input**: Image I, **Output**: Enhanced Image O
TS = 8, N = 0, LEI = I, Clipping_Factor = 3
GEI = GHE(LEI)
Prev_F = Evaluate_Similarity(GEI, LEI)
While N < (TS * TS − 1):
      GEI = GHE(LEI)
      New_LEI = CLAHE(Cipping_Factor, LEI)
      F = Evaluate_Similarity(GEI, New_LEI)
      If F > Prev_F:
            Prev_F = F
            LEI = New_LEI
            Clipping_Factor + = 1
      Else:
            Return (LEI, Clipping_Factor − 1)
      N + = 1
Return (LEI, N − 1)

---

value. The clipping factor is initially set to 3, and during the iterations, we intend to find the best value. Let LEI denote the Locally Enhanced Image that is initially set to the input image $I$. In each iteration, we run the CLAHE algorithm with the current clipping factor and the LEI to obtain a new LEI. Also, we compute the GHE of the new LEI. Then, we evaluate the similarity between the new LEI and the GEI using a similarity evaluation metric (for example, SSIM). If the result is better than in the previous iteration, it means more global characteristics are preserved. Hence, we update the LEI and run CLAHE with a greater clipping factor. If a lower quantity in terms of similarity (SSIM) is obtained, the algorithm is stopped and returns the previous result.

## 5. Experimental Results

In this section, we describe the dataset that has been employed in our experiments. Several key similarity/dissimilarity metrics are used to compare our algorithm with other approaches. A large dataset of chest X-ray images[5] is used; among them, 1349 images are used for the training. These images have different sizes from $912 \times 672$ to $2916 \times 2663$ pixels. For the experiments, images are selected randomly to ensure the results are independent of a particular set of images.

Unlike similarity metrics such as PSNR or SSIM, which are used to assess the similarity between two images, we employ the evaluation metrics focused on specific characteristics of the enhanced images to quantitatively evaluate our proposed approach. While the similarity metrics are useful for assessing the similarity between two images, they do not fully capture the effectiveness of an image enhancement algorithm. For example, a high PSNR or SSIM value does not necessarily indicate that the enhanced image is more suitable for its intended application or that it preserves important features such as edges or contrast. Therefore, our evaluation metrics delve deeper into these aspects to provide a more comprehensive assessment of the algorithm's performance.

These evaluation metrics include the following:

(1) Edge Count: The total number of edges detected in the enhanced image using the Canny edge detection algorithm. A higher edge count may signify enhanced image clarity and contrast because it means more edges in the image are detectable after enhancement.

(2) Edge Density: The density of edges in the enhanced image is calculated as the ratio of edge pixels to total pixels. A higher edge density suggests a more pronounced presence of distinguishing features.

(3) Entropy: It is a measure of randomness or uncertainty in the intensity values of the enhanced image. A lower entropy value indicates a more predictable and structured pixel distribution.

(4) Mean Value: It is the average intensity value of the pixels in the enhanced image. A higher mean value indicates a brighter image.

(5) Average Gradient: It is the average magnitude of the gradient of intensity values across the image. A higher average gradient value indicates a more pronounced transition between intensity levels, resulting in a sharper image feature.

These metrics provide valuable insights into how well our algorithm can detect edges, enhance contrast, and maintain important image characteristics. By quantitatively analyzing these aspects, we can better understand the performance of our algorithm and compare it with existing methods in the literature.

We have used the mentioned evaluation metrics to compare our algorithm with other methods. However, we have also experimented with many different similarity metrics. Structural Similarity Index (SSIM) is a widely used similarity metric for assessing the similarity between two images. Unlike PSNR, which considers pixel-wise differences, the SSIM computes the similarity based on structural information, luminance, and contrast of the images, and incorporates characteristics of the human visual system. It is the reason that we prefer using SSIM over other measures as the evaluation metric. The SSIM's resulting value is within the range $[-1, 1]$. Table 3 describes what different SSIM values mean.

We also employ Peak Signal-to-Noise Ratio (PSNR) as a similarity evaluation metric in our experiments. It measures the quality of the reconstructed image compared to the original image by calculating the ratio of the maximum possible power of an image to the power of corrupting noise that affects the fidelity of its representation. The PSNR is expressed in decibels (dB), and a higher value indicates better image quality.

Table 3.   Description of SSIM values.

| Value | $-1$ | 0 | 1 |
|---|---|---|---|
| Description | Totally dissimilar images | Not similar images | Totally similar images |

The Mean Squared Error (MSE) is another commonly used metric to measure the difference between two images. It calculates the average of squares of the differences between the corresponding pixels of two images. A lower MSE value indicates a higher similarity between images. Hence, it is basically a dissimilarity measure, and in each iteration, having it decreased compared to the previous iteration means improvement.

The Structural Content Index (SCI) is a metric that quantifies the difference in content between two images based on their structural information. It evaluates the preservation of the structural features of the original image in the enhanced image.

The Root Mean Square Error (RMSE) is similar to MSE but takes the square root of the average squared differences between corresponding pixels of the images. It provides a measure of the average magnitude of the errors between predicted and observed values. It can be basically used as a dissimilarity metric.

The Mean Absolute Error (MAE) measures the average absolute differences between corresponding pixels of the images. It is also a dissimilarity metric.

All of above similarity/dissimilarity metrics can help us quantitatively assess the iterations/steps of G-CLAHE, so that we can decide when to terminate the algorithm. In one experiment, we select 100 random medical X-ray images and apply the G-CLAHE algorithm with all the mentioned similarity metrics individually. The results of the detected edges and edge densities are shown in Table 4, where all the similarity metrics are pretty close. This aligns with the fact that our proposed method is completely independent of the chosen similarity metric. On the other hand, by comparing the edge density precisely, we can observe that the G-CLAHE with SSIM is able to detect a few more edges.

For comparisons, we have implemented the TV-Homomorphic approach,[16] PLIP_Unsharp_Masking,[18] Multiscale_Retinex,[4,12] GHE, CLAHE, and G-CLAHE. We compare their corresponding results visually as well. Note that the tile size is fixed as 8, and the clipping factor for CLAHE is set to 2. Figure 7 shows two random X-ray sample images being operated by GHE, CLAHE, and G-CLAHE separately. In the first row, G-CLAHE achieves SSIM = 0.95, and the best clipping factor 11 is obtained. In the second row, G-CLAHE achieves SSIM = 0.94, and the best clipping factor 28 is obtained. It is obvious that the result of G-CLAHE has the best contrast representation and the details are more distinguishable with G-CLAHE. Also, CLAHE has shown a better quality compared to GHE.

Table 4.   Results of G-CLAHE using different similarity metrics.

|  | SSIM | PSNR | MSE | SCI | RMSE | MAE |
|---|---|---|---|---|---|---|
| Edge Count | 305320.60 | 303525.80 | 303310.50 | 302785.90 | 303310.50 | 302534.40 |
| Edge Density | 0.1203 | 0.1195 | 0.1194 | 0.1193 | 0.1194 | 0.1191 |

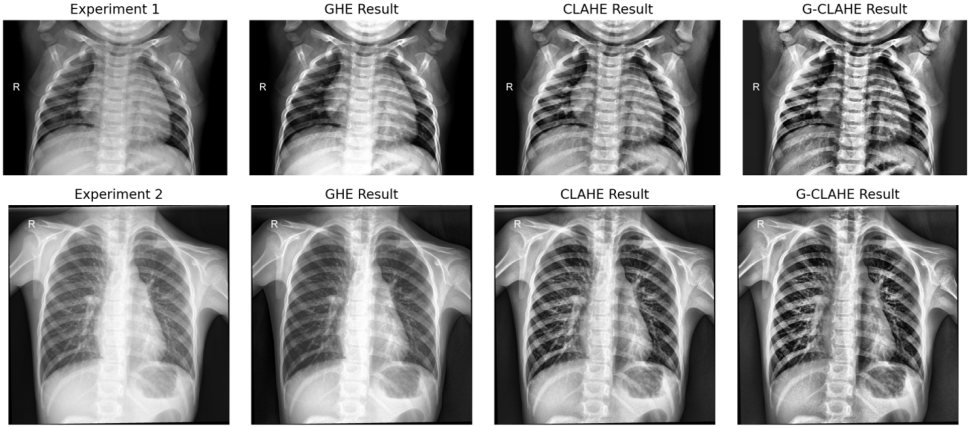

Fig. 7.  Applying GHE, CLAHE, and G-CLAHE on two sample X-ray images.

In Fig. 8, another X-ray image is operated by TV-Homomorphic, PLIP_Unsharp_ Masking, Multiscale_Retinex, and G-CLAHE. The G-CLAHE achieves SSIM = 0.97, and the best clipping factor 10 is obtained. The enhanced images by TV-Homomorphic and G-CLAHE overcome the other methods in terms of the clarity of the details. Specifically, G-CLAHE is the best approach in terms of contrast enhancement since it shows some very detailed parts of the image that do not belong to the main obvious components.

Our qualitative experiments show that G-CLAHE overcomes the other methods. We apply Canny Edge Detection to compute the edge matrices of the resulting image of each algorithm. Then, we count edge density and how many edges can be detected by applying each algorithm. These quantitative evaluation metrics can help us to compare G-CLAHE with other methods. Table 5 shows the results by averaging over 100 random X-ray images in the dataset. In this experiment, the tile size is 8 for both CLAHE and G-CLAHE. Also, the clipping factor is set to 2 for CLAHE. The average value of the clipping factors selected by G-CLAHE is 19. Also, the mean SSIM is 0.92.

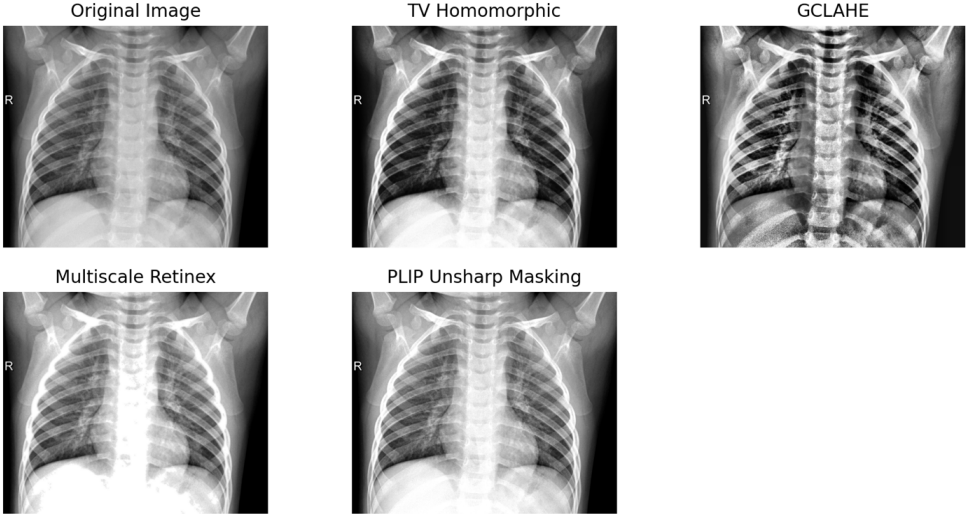

Fig. 8.  Applying TV-Homomorphic, G-CLAHE, Multiscale_Retinex, and PLIP Unsharp Masking on a sample X-ray image.

Table 5.   Quantitative comparison of the average results of G-CLAHE with other methods.

|  | Edge Count | Edge Density | Mean Value | Entropy | Average Gradient |
|---|---|---|---|---|---|
| Original Image | 4312.30 | 0.0020 | 119.95 | 7.4337 | 21.83 |
| GHE | 18586.10 | 0.0073 | 119.43 | 7.2869 | 30.24 |
| CLAHE | 40248.60 | 0.0153 | 127.20 | 7.6363 | 39.33 |
| G-CLAHE | 327681.70 | 0.1161 | 120.70 | 7.7622 | 61.23 |
| TV-Homomorphic | 12232.00 | 0.0146 | 119.94 | 5.1129 | 33.49 |
| Multiscale_Retinex | 3556.00 | 0.0170 | 139.75 | 7.4131 | 34.59 |
| PLIP Unsharp Masking | 5391.00 | 0.0064 | 130.61 | 7.4316 | 30.22 |

It can be inferred from Fig. 9 that the edge density and the number of detected edges in G-CLAHE are significantly higher compared to the other methods. Figure 10 compares different methods in terms of the mean value of the pixels. An interesting point about the G-CLAHE algorithm is that it has almost kept the average brightness level of the original image. This can be understood by the Mean value column. TV-Homomorphic and GHE have also achieved this characteristic. TV-Homomorphic has resulted in the minimum entropy among all the methods. On the other hand, the average gradient of G-CLAHE is significantly higher than the other methods, indicating that it can potentially preserve more detailed information and enhance image features effectively.

Having proven the strength of G-CLAHE over other methods, we describe our experimental results regarding the key parameter in G-CLAHE. Although G-CLAHE finds the best clipping factor automatically, tile size remains almost the only key parameter in G-CLAHE that needs to be chosen wisely.

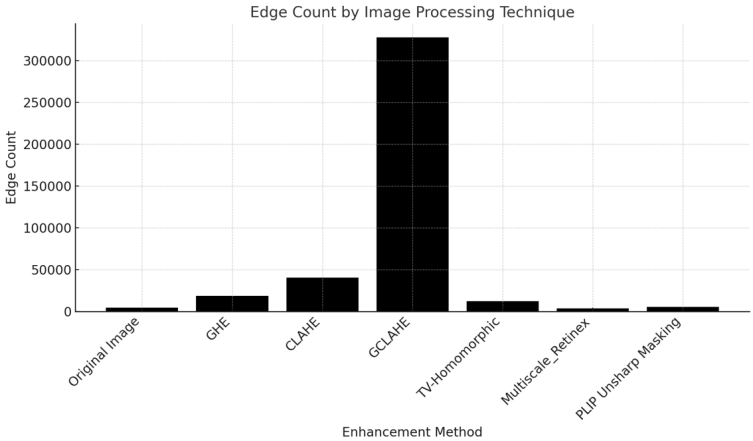

Fig. 9.   Number of detected edges by the Canny method for different enhancement algorithms.

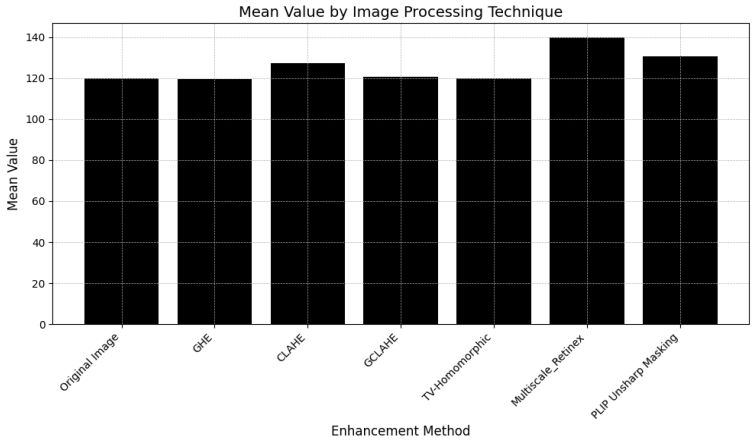

Fig. 10.   The mean value of pixels for different enhancement algorithms.

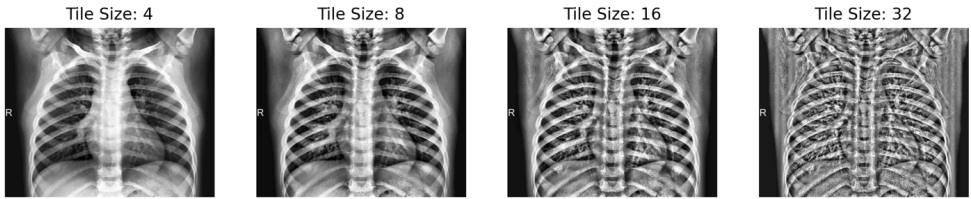

Fig. 11.   G-CLAHE algorithm with different tile sizes.

We run G-CLAHE with sample inputs and different tile sizes to compare how tile size can affect the qualitative features of the original image. Figure 11 shows the result of applying the G-CLAHE algorithm with different tile sizes. It is observed that having a very small tile size (tile_size = 4) is not very effective and cannot increase the contrasts of the image properly. On the other hand, having a very large tile size (tile_size = 32) causes noise over-amplification and generates too much contrast. Hence, the tile size should be within an appropriate range, and such range for this specific dataset is approximately from 8 to 16.

## 6.  Conclusions

In summary, medical X-ray images are crucial for medical diagnoses, but they often have low contrast. The development of G-CLAHE offers a new way to enhance these images, specifically tailored for X-rays. By combining the strengths of GHE and CLAHE, G-CLAHE preserves both local and global image characteristics while avoiding common issues like noise over-amplification. The experiments have shown that our framework does not depend on any specific similarity metric. Through evaluation, G-CLAHE has proven to be highly effective for enhancing X-ray images, offering a promising solution for clearer and more accurate medical diagnoses.

## References

1. A. Bakar, K. Muchtar, F. Arnia, T. Y. Arif, H. Kurniawan and N. Maulina, Image enhancement effects on chest X-ray images for COVID-19 detection, in *2023 3rd Int. Conf. Computing and Information Technology*, 13-14 September 2003, Tabuk, Saudi Arabia, pp. 243–248.
2. C. H. Bindu and B. S. Chandra, Medical images enhancement by homomorphic filtering equalization, *Int. J. Image Process. Appl.* **3**(7) (2016) 183–185.
3. H. Ibrahim and N. S. P. Kong, Brightness preserving dynamic histogram equalization for image contrast enhancement, *IEEE Trans. Consum. Electron.* **53**(4) (2007) 1752–I758.
4. I. Belykh, Homomorphic filtering for radiographic image contrast enhancement and artifacts elimination, in *Proc. 3rd Int. Conf. on Frontiers of Intelligent Computing: Theory and Applications* (Springer, Cham, 2014), pp. 423–430.
5. D. Kermany, K. Zhang and M. Goldbaum, Large dataset of labeled optical coherence tomography (OCT) and chest X-Ray images, *Mendeley Data*, Version 3 (2018).
6. M. Kim and M. G. Chung, Recursively separated and weighted histogram equalization for brightness preservation and contrast enhancement, *IEEE Trans. Consum. Electron.* **54**(3) (2008) 1389–1397.
7. T. Kim and J. Paik, Adaptive contrast enhancement using gain-controllable clipped histogram equalization, *IEEE Trans. Consum. Electron.* **54**(4) (2008) 1803–1810.
8. K. K. Lavania and R. Kumar, Image enhancement using filtering techniques, *Int. J. Comput. Sci. Eng.* **4**(1) (2012) 14–20.
9. S. M. Pizer *et al.*, Adaptive histogram equalization and its variations, *Comput. Vis. Graph. Image Process.* **39**(3) (1987) 355–368.
10. S. M. Pizer, R. E. Johnston, J. P. Ericksen, B. C. Yankaskas and K. E. Muller, Contrast-limited adaptive histogram equalization: Speed and effectiveness, in *Proc. First Conf. on Visualization in Biomedical Computing*, 22–25 May 1990, Atlanta, GA, pp. 337–345.
11. A. W. Setiawan, Effect of chest X-Ray contrast image enhancement on pneumonia detection using convolutional neural networks, in *2021 IEEE Int. Biomedical Instrumentation and Technology Conf.*, 20-21 October 2021, Yogyakarta, Indonesia, pp. 142–147.
12. S. Setty, N. K. Srinath and M. C. Hanumantharaju, Development of multiscale retinex algorithm for medical image enhancement based on multi-rate sampling, in *Proc. Int. Conf. on Signal Processing, Image Processing & Pattern Recognition*, 7-8 February 2013, Coimbatore, India, pp. 145–150.
13. M. Sharma and D. Kumar, Comparative analysis of image enhancement techniques for chest X-ray images, in *2022 Int. Conf. on Computational Intelligence and Sustainable Engineering Solutions*, 20-21 May 2022, Greater Noida, India, pp. 130–135.
14. M. H. Siddiqi, X-Ray image enhancement using global histogram equalization, in *Proc. 29th Int. Conf. on Computer Theory and Applications*, 29-31 October 2019, Alexandria, Egypt, pp. 90–95.
15. S. Sukhjinder, R. K. Bansal and S. Bansal, Medical image enhancement using histogram processing techniques followed by median filter, *Int. J. Image Process. Appl.* **3**(1) (2012) 1–9.

16. R. Wang and G. Wang, Medical X-ray image enhancement method based on TV-homomorphic filter, in *Proc. 2nd Int. Conf. on Image, Vision and Computing*, 2-4 June 2017, Chengdu, China, pp. 315–318.

17. G. Yadav, S. Maheshwari and A. Agarwal, Contrast limited adaptive histogram equalization based enhancement for real time video system, in *Proc. Int. Conf. on Advances in Computing, Communications and Informatics*, 24–27 September 2014, Delhi, India, pp. 2392–2397.

18. Z. Zhao and Y. Zhou, PLIP based unsharp masking for medical image enhancement, in *Proc. IEEE Int. Conf. on Acoustics, Speech and Signal Processing*, 20–25 March 2016, Shanghai, China, pp. 1238–1242.

19. K. Zuiderveld, Contrast limited adaptive histogram equalization, in *Graphics Gems lV*, ed. P. Heckbert (Academic Press, 1994).

---

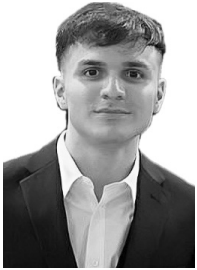

**Sohrab Namazi Nia** received his B.Sc. in Computer Engineering from the Iran University of Science and Technology (IUST), Tehran, Iran, in 2022. He graduated with a concentration in Artificial Intelligence and was awarded the best computer engineering student of IUST in 2019. He achieved a GPA of 4.0/4.0 at IUST. Sohrab was also a member of the National Organization for Development of Exceptional Talents (SAMPAD) for seven years. He was also ranked among the top 0.4% of the Iran University Entrance Exam (Konkour)

Sohrab has extensive experience in both academia and industry. He has served as a teaching assistant and instructor 13 times during his undergraduate and Ph.D. studies. Professionally, he has worked as an ML intern and full-stack software engineer at the Tiva System, Tehran, Iran. His projects span artificial intelligence and software engineering, focusing on machine learning, deep learning, reinforcement learning, and backend development using ASP.

Net Core. Some projects include policy reusability in RL, facial emotion recognition using neural networks, Eventus (Web app), and GoNuts (Android App). His work can be found on his GitHub profile.

Currently, Sohrab is direct-Ph.D. candidate at the New Jersey Institute of Technology (NJIT) and a research assistant at the NJIT Big Data Analytics Lab (BDaL). His research focuses on optimizing machine learning and graph search models for efficiency. He explores optimization opportunities arising from man-machine collaboration and addresses data management and computational challenges to enable large-scale analytics with humans in the loop. He has worked on various projects, including efficient policy reusability in RL and answering lower-bound distance queries in large-scale real-world graphs. His current research interests include Machine Learning, Deep Learning, Reinforcement Learning, Deep RL, Computer Vision and Image Processing, Optimization of Graph Search Algorithms, Natural Language Processing, and Software Engineering.

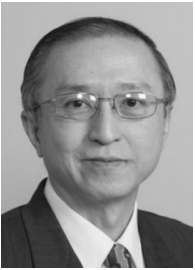

**Frank Y. Shih** received B.S. from the National Cheng Kung University, Tainan, Taiwan, in 1980, M.S. from the State University of New York, Stony Brook, USA, in 1984, and Ph.D. from the Purdue University, West Lafayette, Indiana, USA, in 1987. He is a Professor jointly appointed in the Department of Computer Science, the Department of Electrical and Computer Engineering, and the Department of Biomedical Engineering at the New Jersey Institute of Technology, Newark, New Jersey. He currently serves as the Director of Artificial Intelligence and Computer Vision Laboratory.

Dr. Shih held a visiting professor position at the Princeton University, Columbia University, National Taiwan University, National Institute of Informatics, Tokyo, Conservatoire National Des Arts Et Metiers, Paris, and Nanjing University of Information Science and Technology, China. He is an internationally renowned scholar and currently serves as Editor-in-Chief for the *International Journal of Pattern Recognition and Artificial Intelligence*. He was Editor-in-Chief for the *International Journal of Multimedia Intelligence and Security*. In addition, he is on the Editorial Board of 12 international journals. He has served as a steering member, session chair, and committee member for numerous professional conferences and workshops. He has received numerous grants from National Science Foundation, NIH, NASA, Navy and Air Force, and Industry. He has won the Research Initiation Award from NSF, the Outstanding Teaching Award and the Board of Overseers Excellence in Research Award from NJIT, and the Best Paper Awards from journals and conferences.

Dr. Shih is internationally recognized as an expert in Artificial Intelligence and Pattern Recognition, Deep Learning, Watermarking, Steganography, and Forensics. He has authored 6 books including "Digital Watermarking and Steganography", "Image Processing and Mathematical Morphology", "Image Processing and Pattern Recognition", and "Multimedia Security: Watermarking, Steganography, and Forensics". He has published over 154 journal papers, 109 conference papers, and 23 book chapters. His current research interests include artificial intelligence, deep learning, image processing, watermarking and steganography, digital forensics, pattern recognition, bioinformatics, biomedical engineering, fuzzy logic, and neural networks.